\documentclass[conference]{IEEEtran}
\IEEEoverridecommandlockouts
\usepackage{cite}
\usepackage{amsmath,amssymb,amsfonts}
\usepackage{graphicx}
\usepackage{textcomp}
\usepackage{xcolor}
\def\BibTeX{{\rm B\kern-.05em{\sc i\kern-.025em b}\kern-.08em
    T\kern-.1667em\lower.7ex\hbox{E}\kern-.125emX}}

\usepackage{graphicx}
\usepackage{color,hyperref}
\usepackage{amsmath,amssymb}
\usepackage{amsmath}
\usepackage{lipsum}
\usepackage{algorithmicx}
\usepackage{subcaption}
\usepackage{comment}
\usepackage{url}
\usepackage{multirow}
\usepackage{adjustbox}
\DeclareUnicodeCharacter{2212}{-}

\begin{document}

\title{\textit{Does Yoga Make You Happy?} Analyzing Twitter User Happiness using Textual and Temporal Information}
\author{\IEEEauthorblockN{Tunazzina Islam}
\IEEEauthorblockA{\textit{Department of Computer Science} \\
\textit{Purdue University}\\
West Lafayette, IN, USA \\
islam32@purdue.edu}
\and
\IEEEauthorblockN{Dan Goldwasser}
\IEEEauthorblockA{\textit{Department of Computer Science} \\
\textit{Purdue University}\\
West Lafayette, IN, USA  \\
dgoldwas@purdue.edu}
}

\maketitle

\begin{abstract}
Although yoga is a multi-component practice to hone the body and mind and be known to reduce anxiety and depression, there is still a gap in understanding people's emotional state related to \textit{yoga} in social media. In this study, we investigate the causal relationship between \textit{practicing yoga} and \textit{being happy} by incorporating textual and temporal information of users using Granger causality. To find out causal features from the text, we measure two variables (i) Yoga activity level based on content analysis and (ii) Happiness level based on emotional state. To understand users' yoga activity, we propose a joint embedding model based on the fusion of neural networks with attention mechanism by leveraging users' social and textual information. For measuring the emotional state of yoga users (target domain), we suggest a transfer learning approach to transfer knowledge from an attention-based neural network model trained on a source domain. Our experiment on Twitter dataset demonstrates that there are $1447$ users where ``yoga Granger-causes happiness".
\end{abstract}


\section{Introduction}
Yoga focuses on the body-mind-spirit connection and its importance in all aspects of personal development and well-being by incorporating a wide variety of postural/exercise, breathing, and meditation techniques \cite{goyeche1979yoga}. 
Many studies show that yoga can alleviate symptoms of anxiety and depression as well as promote good physical fitness \cite{khalsa2004treatment, yurtkuran2007modified, smith2009evidence, ross2010health}. The main reason for yoga’s growing popularity \cite{birdee2008characteristics} is the large-scale transmission of education. Interest in this topic can come from either practitioners or commercial parties. Despite the current popularity of yoga, there is little research on analyzing people's life-style decision about yoga in social media.

\begin{figure*}
\begin{subfigure}{.5\textwidth}
  \centering
  \includegraphics[width=\textwidth]{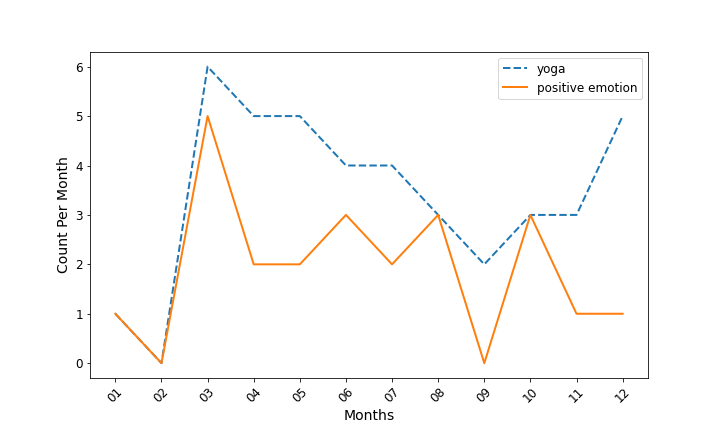}
  \caption{Year 2018}
  \label{fig:y2018}
\end{subfigure}
\begin{subfigure}{.5\textwidth}
  \centering
  \includegraphics[width=\textwidth]{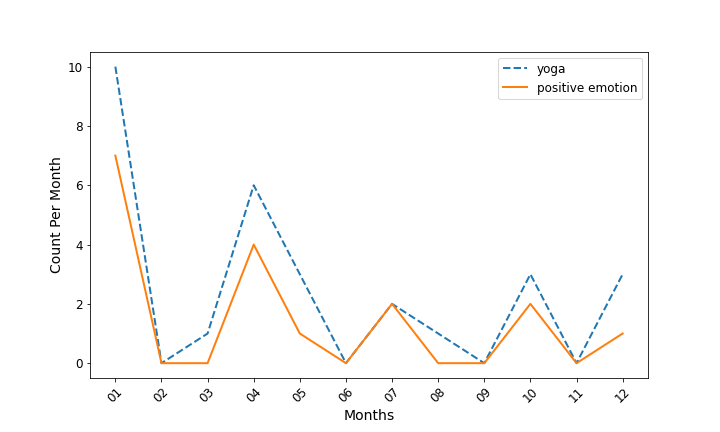}
  \caption{Year 2019.}
  \label{fig:y2019}
\end{subfigure}
\caption{Example of causal features of a practitioner in two different years. We have two causal features. One is Yoga activity (blue dotted line) and another is Positive emotion (orange solid line).}
\label{fig:GC_feature}
\end{figure*}
Nowadays people spend a significant amount of time on social media to express opinions, interact with friends, share ideas and thoughts. As a result, social media data contain rich information providing opportunities to study language and social behavior on a large scale. Leveraging the social media data effectively have been increasingly used to better understand human minds and predict human behavior. Prior research has demonstrated that by analyzing the information in a user's social media account, we can infer many latent user characteristics like personality \cite{kosinski2013private, schwartz2013personality}, emotions \cite{wang2015detecting}, mental health \cite{amir2017quantifying}, mental disorders \cite{de2013predicting, reece2017forecasting}.

Over the past several years, many studies have been devoted towards exploring individual well-being on social media \cite{schwartz2016predicting, schwartz2013characterizing, yang2016life, islam2019yoga}. However, finding causal link between yoga and happiness through the language people use on social media remains unexplored. 

In this paper, we find causal relationship between \textit{doing yoga} and \textit{being happy} by incorporating textual and temporal information of practitioners using Granger causality \cite{granger1988some}. We use well-being related tweets from Twitter focusing on \textit{Yoga}. 


We measure two variables to extract causal features from text. One is yoga activity level based on tweet content analysis and another is happiness level based on the emotional state of practitioner. Fig. \ref{fig:GC_feature} shows the example of causal features of a practitioner from our dataset in year $2018$ and $2019$ where we notice a trend between doing yoga and having positive emotion. The causal feature temporally causes rise of positive emotion of the user in March 2018 (Fig. \ref{fig:y2018}) and April 2019 (Fig. \ref{fig:y2019}).

To measure yoga activity level, we predict user type who tweets about yoga whether they are practitioner, promotional, or others i.e. appreciate yoga by tweeting/retweeting about yoga but do not practice yoga. In this work, we propose a method for combining a large amount of Twitter content and social information associated with each user by building a joint embedding attention-based neural network model called Yoga User Network (YUN). 


For emotion detection task, we build an attention-based neural network model exploiting the labelled emotion corpus (source domain) collected by reference \cite{saravia2018carer}. As our source task and target task are same, we employ transfer learning approach \cite{pan2009survey} to detect emotion of yoga practitioners (target domain) focusing on Ekman’s \cite{ekman1992argument} 6 basic emotions (skipping \textit{disgust} but adding \textit{love}). We compare our emotion detection model with baseline model which is gated recurrent neural network (GRU) \cite{cho2014properties} model. For sanity check, we randomly sample $550$ tweets and label them manually with 6 emotions \textit{\{joy, love, sadness, anger, fear, surprise\}} as well as including no emotion \textit{\{ne\}} tag and calculate accuracy and macro average $F1$-score. 

The main contributions of this paper are as follows:
\begin{enumerate}
    
    \item We suggest a joint embedding attention-based neural network model called YUN that explicitly learns user embedding leveraging tweet text, emoji, metadata, and user network to classify user type.
    
     \item We evaluate our user type detection model under ten different feature settings: (i) Description; (ii) Location; (iii) Tweets; (iv) Network; (v) BERT (Bidirectional Encoder Representations from Transformers)\cite{devlin2018bert} fine-tuned with Description (Description\_BERT); (vi) BERT fine-tuned with Location (Location\_BERT); (vii) BERT fine-tuned with Tweets (Tweets\_BERT); (viii) joint embedding on description and location (Des + Loc); (ix) joint embedding on description, location, and tweets (Des + Loc + Twt); (x) joint embedding on description, location, and network (Des + Loc + Net). We show that YUN outperforms these settings.
     
    \item We propose an attention-based Bi-directional LSTM model for emotion detection of source domain. We employ transfer learning for emotion detection of yoga practitioners (target domain). 
    
    \item We show that our Bi-LSTM attention model outperforms the baseline based on the sanity check.
    
     \item We analyze Granger causality between yoga activity and happiness on Twitter data. Among $8813$ practitioners who tweeted about `yoga' and having positive emotion, we observe that for $1447$ practitioners ``yoga Granger-causes happiness". 
     
\end{enumerate}

\textbf{Reproducibility:} Our code and the data are available at this link \href{https://github.com/tunazislam/Causal-Yoga-Happiness}{Causal-Yoga-Happiness}.\footnote[1]{\url{https://github.com/tunazislam/Causal-Yoga-Happiness}}

The rest of the paper is organized as follows:
\hyperref[sec:2]{Section II} discusses the relevant literature; \hyperref[sec:3]{Section III} provides problem definition of our work; next, \hyperref[sec:4]{Section IV} describes the proposed methodology; then, \hyperref[sec:5]{Section V} has details of dataset; later, \hyperref[sec:6]{Section VI} discusses the baseline models, hyperparameter tuning, and experimental results.

\section{Related Work}
\label{sec:2}
Prior works on causality detection \cite{qiu2012granger, acharya2014causal, kang2017detecting} in time series data use Granger causality for predicting future values of a time series using past values of its own and another time series. 
Our study focuses on the reason to see whether yoga makes people happy by employing Granger causality test using textual and temporal information of the user. 

Research on understanding the demographic properties of Twitter users by exploiting user's community information along with textual features \cite{li2015learning, benton2016learning, yang2017overcoming, mishra2018neural, del2019you} has become increasingly popular. Our main challenge is to use these information to construct a coherent user representation relevant for the specific set of choices that we are interested in. We propose a joint embedding attention-based neural network model which explicitly learns Twitter user representations by leveraging social and textual information of users to understand their types i.e. practitioner, promotional, and others.

Besides, emotion is a key aspect of human life and have a wide array of applications in health and well-being. Early works on emotion detection \cite{alm2005emotions, strapparava2007semeval} focused on
conceptualizing emotions by following Ekman’s model of six basic emotions: anger, disgust, fear, joy, sadness and surprise \cite{ekman1992argument}. Reference \cite{wang2012harnessing} collected a large emotion corpus ($5$ millions) for $5$ of
Ekman’s $6$ basic emotions (skipping \textit{disgust}), but
adding \textit{love} and \textit{thankfulness}. References
\cite{suttles2013distant, meo2017processing}, and \cite{abdul2017emonet} followed the Wheel of Emotion \cite{plutchik1980general} considering emotions as a discrete set of eight basic emotions. Emotion recognition of yoga user still suffers from the bottleneck of labeled data. So we use transfer learning approach to detect emotion of yoga practitioners focusing on six emotions \textit{\{joy, love, sadness, anger, fear, surprise\}}.



\section{Problem Statement}
\label{sec:3}
In this section, we define two major parts of our work: A) Find Granger causality, B) Transfer knowledge from source to target.  

\subsection{Granger Causality}
\label{subsec:3.1}
Granger causality \cite{granger1988some} is a probabilistic account of causality method to investigate causality between two variables in a time series. It is developed in the field of econometric time series analysis. Granger formulated a statistical definition of causality based on the essential assumptions that (i) a cause occurs before its effect and (ii) 
knowledge of a cause can be used to predict its effect. A time series $X$ (source) is said to Granger-cause a time series $Y$ (target) if past values $x_{t-i}$ are significant indicators in predicting $y_t$. So future target value $y_t$ depends on both past target time series $y_{t−i}$ and past source time series $x_{t-i}$ rather than only past target time series $y_{t-i}$.

Let's assume, we compute the Yoga activity level $(a)$ and Happiness level $(p)$. According to Granger causality, given a target time series $p$ (effect) and a source time series $a$ (cause), predicting future target value $p_t$ depends on both past target $(p_{< t})$ and past source time series $(a_{< t})$. We calculate Granger causality as follows:
\begin{align}
GC(p_t | p_{< t}, a_{< t}) = \sum_{i=1}^m \alpha_i p_{t-i} + \sum_{j=1}^n \beta_j a_{t-j} \label{eq:1}
\end{align}
where, $m$ and $n$ are size of lags in the past observation, $\alpha$ and $\beta$ are learnable parameters to maximize the prediction expectation.

\begin{figure*}[htbp]
  \centering  
  \includegraphics[width= 0.8 \textwidth]{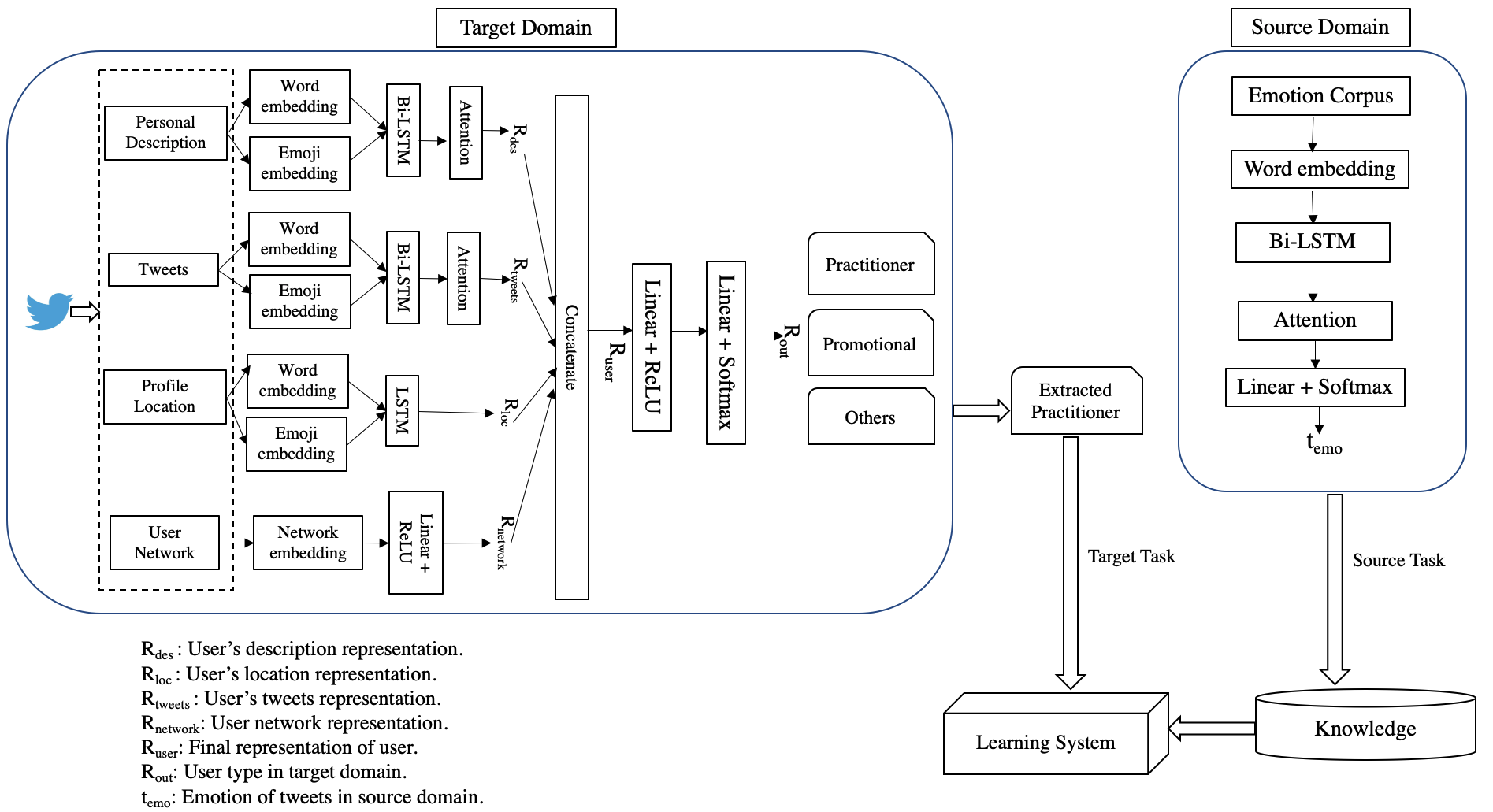}
    \caption{Overall architecture of our work. Inside the target domain (blue box of left-hand side), we have YUN model. YUN has four sub-networks: Description, Location, Tweets, and User Network. Description and Tweets sub-networks use Attention based Bi-LSTM network to generate representation of description, $R_{des}$ and tweets, $R_{tweets}$ respectively. Location sub-network uses only LSTM to create location representation, $R_{loc}$ because the length of sequence for location is shorter. User Network uses linear layer with ReLU activation to compute the representation of user network, $R_{network}$. Concatenation of these four representations generate the final user representation, $R_{user}$ which is fed into a fully connected two-layer classifier activated by ReLU and softmax respectively. Inside the source domain (blue box of right-hand side), we have attention-based Bi-LSTM emotion detection model to classify emotion ($t_{emo}$) of source domain. Transfer learning extracts the knowledge from the source task and
    applies the knowledge to the target task (classify the emotion of yoga practitioners). }
    \label{fig:model}
\end{figure*}

\subsection{Transfer Knowledge}
\label{subsec:3.2}
In our work, given a source domain $D_{src}$ and learning task $T_{src}$, a target domain $D_{tgt}$ and learning task
$T_{tgt}$, transfer learning aims to help improve the learning of the target predictive function $f_T(.)$ in $D_{tgt}$ using the knowledge in $D_{src}$ and  $T_{src}$, where  $D_{src} \ne D_{tgt} $ and $T_{src} = T_{tgt} $.

\subsubsection{Target domain}
\label{subsubsec:3.2.1}
Our target domain consists of yoga practitioners information. We extract the practitioners using YUN model which employs description, location, yoga-related tweets of users, and user network and then jointly builds a neural network model to generate a dense vector representation for each field and finally concatenates these representations as to the feature for multi-class classification. 

To obtain the ground-truth of YUN model, we manually annotate by labeling practitioner as `0', promotional as `1', others as `2' (annotation details in \hyperref[subsec:5.1]{subsection V.A}). 

In YUN model, each feature is processed by a separate sub-network to generate a feature vector representation $R_i$. The output of each sub-networks are feature vectors like $ R_1 \in \mathbb{R}^{d^1}$, $ R_2 \in \mathbb{R}^{d^2}$ $,....,$ and $ R_N \in \mathbb{R}^{d^N}$ respectively, where $N$ is the number of sub-networks. These feature vectors are then concatenated to build a final user representation $R_{user}$ which is fed into a two-layer classifier consisting of a layer $L_{user} \in \mathbb{R}^{({d^1}+{d^2}+....+{d^N}) \times h}$, where $h$ is a model parameter, and a layer $L_{out} \in \mathbb{R}^{h \times l}$, where $l$ is the number of output classes. The final output, $R_{out}$ is fully connected to the output layer and activated by softmax to generate a probability distribution over the classes. The final prediction is computed as follows:
\begin{align}
R_{out} = softmax(W_{out} Z_{user} + b_{out}) \label{eq:2}
\end{align}
where,
\begin{align}
Z_{user} &= relu (W_{user} R_{user} + b_{user}), \label{eq:3} \\
R_{user} &= R_1 || R_2 .... || R_N , \label{eq:4}
\end{align}
$W_{user}$ \& $W_{out}$ are weights and $b_{user}$ \& $b_{out}$ are biases. 
We denote the concatenation operation as $||$.

We use \textit{cross-entropy loss} as the objective function. Let say, $U$ is the number of users and $l$ is the number of class label, then the \textit{cross-entropy loss} is as following:
\begin{align}
CE =  - \sum_{i=1}^U \sum_{j = 1}^l y_i^j log(\hat{y}_i^j)\label{eq:5}
\end{align}
where $y_i$ is the ground-truth of user type and $\hat{y}_i$ is the predicted probability vector. So, $\hat{y}_i^j$ is the probability that user $i$ is a $j$ type of user. We minimize the objective function through Gradient Descent with Adadelta \cite{zeiler2012adadelta}.

\subsubsection{Source domain}
\label{subsubsec:3.2.2}
The source domain consists of emotion corpus $E_x$ labeled with 6 emotions \textit{\{joy, love, sadness, anger, fear, surprise\}} where $x$ is total the number of tweets in the corpus. The embeddings of tweets are forwarded to a Bi-LSTM \cite{hochreiter1997long} which creates hidden state $h_i$ at time-step $i$. To assign important words, we use a context-aware attention mechanism \cite{bahdanau2014neural} with a weight, $a_i$ to each word representation. $t_{emo}$ represents the emotion of tweets.

\begin{align}
t_{emo} &= softmax((W_t \times \sum_{i=1}^{x} a_ih_i) + b_t), t_{emo} \in \mathbb{R}^{2D}\label{eq:6}, 
\end{align}
where 
\begin{align}
a_i &= \frac{exp(m_ic_h)}{\sum_{t=1}^x exp(m_tc_h)}\label{eq:7}, \\
m_i &= tanh(W_hh_i + b_h), m_i \in [-1, 1] \label{eq:8},
\end{align}
and with $W_t$ , $W_h$ and $c_h$ are the layer's weights and $b_t$, $b_h$ are the biases. $D$ is the size of each LSTM.

We use \textit{cross-entropy loss} as the objective function. Let say, $x$ is the number of tweets and $e$ is the number of class label, then the \textit{cross-entropy loss} is:
\begin{align}
CE_{emo} =  - \sum_{i=1}^x \sum_{j = 1}^e g_i^j log(\hat{g}_i^j)\label{eq:9}
\end{align}
where $g_i$ is the ground-truth of emotion and $\hat{g}_i$ is the predicted probability vector of emotion. So, $\hat{g}_i^j$ is the probability that tweet $i$ has a $j$ emotion. We minimize the objective function through Gradient Descent with ADAM optimizer \cite{kingma2014adam}.





\section{Technical Approach}
\label{sec:4}
The overall architecture of the proposed model is illustrated in Fig. \ref{fig:model}. Our working pipeline follows four key stages including 1) joint-embedding model (YUN) to classify user type, 2) emotion detection model for source domain, 3) transferring the knowledge of source domain to detect emotions in target domain, 4) finding Granger causality.

\subsection{YUN Model}
\label{subsec:4.1}
YUN is a joint embedding model that uses description, location, tweets, and network for user representation to learn user type.
\subsubsection{Tweet Representation}
\label{subsubsec:4.1.1}
To represent the tweet's textual content, we use pre-trained word embeddings Word2Vec using Skip-gram architecture \cite{mikolov2013efficient}. For emoji embeddings, we use Emoji2Vec \cite{eisner2016emoji2vec}. The embeddings of text and emoji are then forwarded to an attention based Bi-LSTM model to represent user tweet $R_{tweet}$.

\subsubsection{Metadata Representation}
\label{subsubsec:4.1.2}
In this paper, we focus on two metadata fields i.e. user description and location. For description, embeddings of text and emoji are forwarded to Bi-LSTM. The attention mechanism is employed to provide the final representation of user description $R_{des}$  by implementing same approach followed for $R_{tweet}$. 

For user location embedding, embeddings of text and emoji are fed to a LSTM to obtain the representation of user location $R_{loc}\in \mathbb{R}^{D}$, where $D$ is the size of LSTM.

\subsubsection{User Network Construction and Network Representation}
\label{subsubsec:4.1.3}
We construct user network based on $@-$mentions in their
tweets \cite{rahimi2015exploiting}. We create an undirected and unweighted graph from interactions among users via retweets and/or mentions. In this graph, nodes are all users in the dataset (both train and test), as well as other external users mentioned in their yoga-related tweets. An undirected and unweighted edge is created between two users if either user mentioned the other. In this work, we do not consider edge weights. In our data, some users mention themselves in their tweets and we delete those edges to avoid self-loop. Also, some users mention an external user multiple times in their tweets and we consider an undirected, unweighted edge between them.  In
our user network graph, in total, we have $73878$ nodes including all users in the dataset (train and test), as well as other external users mentioned in their tweets and $116085$ edges. 
To compute node embedding, we use Node2Vec \cite{grover2016node2vec}. We generate the embedding of user network and forward to linear layer with ReLU \cite{nair2010rectified} activation function to compute user network representation, $R_{network}$.

\subsubsection{User Representation}
\label{subsubsec:4.1.4}
The final user representation, $R_{user}$ is built by concatenation of the four representations generated from four sub-networks description, location, tweets, and user network respectively. We define $R_{user}$ as follows:

\begin{align}
R_{user} = R_{des} || R_{loc} || R_{tweets} || R_{network} \label{eq:10}
\end{align}
Then $R_{user}$ is passed through a fully connected two-layer classifier. The final prediction $R_{out}$ is computed same as eqn.\eqref{eq:2}.

\subsection{Emotion Detection Model}
\label{subsec:4.2}
Emotions reflect users’ perspectives towards actions and events.  As we do not have ground truth for emotion of our yoga data, we use large emotion corpus ($0.4$ millions) \cite{saravia2018carer} for $5$ of Ekman’s $6$  basic emotions (skipping \textit{disgust}), but adding \textit{love}. 
We use pre-trained word embeddings Word2Vec for tweet embedding and forward the embedding to Bi-LSTM with attention network (described in \hyperref[subsubsec:3.2.2]{III.B.2}). 

Finally, we pass the tweet representation to an one-layer classifier activated by softmax (using eqn.\eqref{eq:6}) to detect emotion of the large emotion corpus $(E_x)$. We refer this model as BiLSTMAttEmo model.

\subsection{Transfer Learning}
\label{subsec:4.3}
We use transfer learning approach by extracting the knowledge from source task (classify emotion of $E_x$ ) and applying the knowledge on target task (classify the emotion of $R_{out}$ if they are yoga practitioners). 

\subsection{Find Causality}
\label{subsec:4.4}
After detecting the practitioner's emotion, we focus two types of emotion \textit{\{joy, love\}} to measure happiness level of each practitioner. 

To measure yoga activity level, we focus on each practitioner's first hand experience based on rule based approach. There are two cases we consider, case (i): Explicitly having First Person Singular Number and First Person Plural Number, case (ii): Implicit first hand experience.
For explicit case, if a tweet contains any of these words \textit {\{``i", ``im",  ``i'm",  ``i've",   ``i'd",  ``i'll",  ``my",  ``me",  ``mine", ``myself", ``we", ``we're",  ``we'd",  ``we'll",  ``we've", ``our", ``ours",  ``us", ``ourselves"\}} with `yoga', we select them as yoga activity. 

For case (ii), there are some tweets which don't explicitly mention First Person Singular Number and First Person Plural Number i.e ``feeling peaceful after doing morning yoga", ``loved yesterday's yoga session \#motivational \#calm". We use parts of speech tagging \cite{toutanova2003feature} to assign parts of speech to each word. We filter out Second and Third Person Singular and Plural Number. We keep those tweets if there are no Verb 3rd-person singular present form (VBZ) and Proper noun singular (NNP) and Noun plural (NNS) and Proper noun plural(NNPS) and Personal pronoun (PRP) and Possessive pronoun (PRP\$) with `yoga' as implicit first hand experience of yoga activity level.

We sort yoga activity level $(a)$ and happiness level $(p)$ increasingly based on DateTime and apply Granger causality (eqn.\eqref{eq:1}) with different lags. We choose lags by running the Granger test several times. Our null hypothesis is ``yoga activity $(a)$  does not Granger-cause happiness $(p)$". We reject the null hypothesis if $p$-value is $\leq 0.05$.


\section{Data}
\label{sec:5}
We download yoga-related tweets using Tweepy. We use Twitter streaming API to extract $419608$ tweets sub-sequentially from May to November, 2019 related to yoga containing specific keywords `yoga', `yogi', `yogalife', `yogalove', `yogainspiration', `yogachallenge', `yogaeverywhere', `yogaeveryday', `yogadaily', `yogaeverydamnday', `yogapractice', `yogapose', `yogalover', `yogajourney'. 
We downloaded timeline tweets of $22702$ users and among them  $15168$ users have at least a yoga-related tweet in their timelines. We have total $35392177$ tweets.

To pre-process the tweets, we first convert them into lower case, remove URLs then use a tweet-specific tokenizer from NLTK to tokenize them. We do not remove emojis.

\begin{figure}[htbp]
  \centering  
  \includegraphics[width= 0.5 \textwidth]{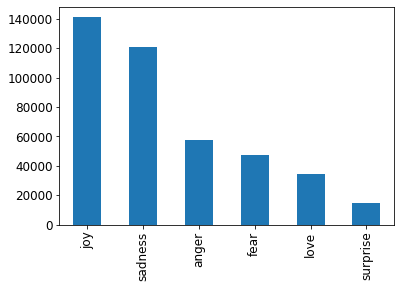}
    \caption{Data distribution of Emotion dataset.}
    \label{fig:emodis}
\end{figure}


\subsection{Data Annotation for YUN Model}
\label{subsec:5.1}
To run YUN model, we need annotated data. We manually annotated $1300$ users (randomly selected from  $15168$ users) based on the intent of the tweets and observation of user description. Consider the following two tweets: 

Tweet 1: Learning some traditional yoga with my good friend. 

Tweet 2: Our mission at 532Yoga is pretty simple; great teachers, great classes and superbly happy students $\#yoga$ 

The intention of Tweet 1 is yoga activity (i.e. learning yoga). Tweet 2 is more about promoting a yoga studio. We annotate user of Tweet 1 as practitioner and user of Tweet 2 as promotional.
\begin{table*}[htbp]
  \centering
  \caption{A summary of hyperparameter settings of the models}
    \begin{tabular}{|c|c|c|c|c|c|c|c|c|c|c|}
    \hline
    \textbf{Model} & \textbf{lr}  & \textbf{opt} & \textbf{reg} & \textbf{word}  & \textbf{emoji}   & \textbf{unit}  & \textbf{attn} & \textbf{h} & \textbf{eut} & \textbf{emo}\\
    \hline
    Description  & 0.01 & Adadelta & 0.0001 & 300 & 300 & 150 & 300 & 200 & 23 & N/A\\
    \hline
    Location & 0.01 & SGD momentum & N/A & 300 & 300 & 150  & 300 & 200 & 9 & N/A\\
    \hline
    Tweets  & 0.01 & Adadelta & 0.0001 & 300 & 300 & 150 & 300 & 200 & 12 & N/A \\
    \hline
    Network  & 0.01 & Adadelta  & 0.0001 & N/A & N/A & N/A & N/A & 200 & 24 & N/A\\
    \hline
    Description\_BERT  & 0.00002 & AdamW & 0.01 & N/A & N/A & N/A & N/A & N/A & 4 & N/A  \\
    \hline
    Location\_BERT  & 0.00002 & AdamW & 0.01 & N/A & N/A & N/A & N/A & N/A & 2 & N/A  \\
    \hline
    Tweets\_BERT  & 0.00002 & AdamW & 0.01 & N/A & N/A & N/A & N/A & N/A & 2 & N/A  \\
    \hline
    Des + Loc  & 0.01 & Adadelta & 0.0001 & 300 & 300 & 150 & 300 & 200 & 14 & N/A\\
    \hline
    Des + Loc + Twt  & 0.01 & Adadelta & 0.0001 & 300 & 300 & 150 & 300 & 200 & 17 & N/A \\
    \hline
    Des + Loc + Net  & 0.01 & Adadelta & 0.0001 & 300 & 300 & 150 & 300 & 200 & 16 & N/A \\
    \hline
    \textbf{YUN}  & 0.01 & Adadelta & 0.0001 & 300 & 300 & 150 & 300 & 200 & 18 & N/A \\
     \hline
    GRUEmo & 0.001 & ADAM & 0.0 & 256 & N/A & 1024 & N/A & N/A & N/A & 10  \\
    \hline
    \textbf{BiLSTMAttEmo}  & 0.01 & ADAM & 0.0001 & 300 & N/A & 150 & 300 & N/A & N/A &21  \\
    \hline
    \hline
    \multicolumn{11}{|p{35em}|}{lr : Learning rate.} \\
    \multicolumn{11}{|p{35em}|}{opt: Optimizer.} \\
    \multicolumn{11}{|p{35em}|}{reg : Weight decay ($L^2$ regularization).} \\
    \multicolumn{11}{|p{35em}|}{word: Word embedding dimension.} \\
    \multicolumn{11}{|p{35em}|}{emoji: Emoji embedding dimension.} \\
    \multicolumn{11}{|p{45em}|}{unit: LSTM unit size for the models except for GRUEmo model (GRU unit size).} \\
    \multicolumn{11}{|p{35em}|}{attn: Attention vector size.} \\
    \multicolumn{11}{|p{35em}|}{h: Size of $L_{user}$ layer which is the first layer of two-layer classifier. } \\
    \multicolumn{11}{|p{35em}|}{eut: Best result achieved at epochs for user type classification. } \\
    \multicolumn{11}{|p{35em}|}{eum: Best result achieved at epochs for practitioner's emotion classification. } \\
     \hline
    \end{tabular}%
  \label{tab:hyperparam}%
\end{table*}%
\subsection{Data Distribution for YUN Model}
\label{subsec:5.2}
In our user type annotated data ($1300$ users), we have $42\%$ practitioner, $21\%$ promotional, and $37\%$ other users. We shuffle the dataset and then split it into train $(60\%)$, validation $(20\%)$ and test $(20\%)$. We use the trained YUN model to find out the user type of remaining $13860$ users.

\subsection{Emotion Dataset}
\label{subsec:5.3}
The emotion dataset has $0.4$ millions tweets annotated with six emotions \textit{\{joy, love, sadness, anger, fear, surprise\}} collected from Twitter using hashtags \cite{saravia2018carer}. The tweets are already pre-processed based on the approach described by reference \cite{saravia2018carer}. The distribution of the data is shown in Fig. \ref{fig:emodis}. We shuffle the dataset and then split it into $60\%$ training, $20\%$ validation and $20\%$ testing data. We use macro average F1-score as the evaluation metric, which is commonly used in emotion recognition studies due to the imbalanced nature of the emotion datasets.

\section{Experimental Settings}
\label{sec:6}

\subsection{Baseline Models}
\label{subsec:6.1}
For emotion detection baseline, to represent texts we initialize word embeddings with index mapping for words. We pass this to gated recurrent neural network (GRU) with batch size= $64$ and then forward to an one-layer classifier activated by softmax. We refer this baseline model as GRUEmo model.

\begin{table*}[htbp]
  \centering
  \caption{Performance comparison of models}
    \begin{tabular}{|c|c|c||c|c|}
          \cline{2-5} \multicolumn{1}{c|}{} & \multicolumn{2}{c||}{\textbf{user type detection}} & \multicolumn{2}{c|}{\textbf{emotion detection}} \\
          \hline
    \textbf{Model} & \textbf{Accuracy}  & \textbf{Macro avg. F1 score}  & \textbf{Accuracy}  & \textbf{Macro avg. F1 score}  \\
    \hline
   Description  & 0.725 & 0.693 & N/A & N/A \\
    \hline
    Location  & 0.676 & 0.563 & N/A & N/A \\
    \hline
    Tweets  & 0.721 & 0.687  & N/A & N/A \\
    \hline
    Network  & 0.752 & 0.557  & N/A & N/A\\
    \hline
    Description\_BERT  &  0.718 & 0.681 &  N/A & N/A \\
    \hline
    Location\_BERT  &   0.679 & 0.606 & N/A & N/A \\
    \hline
    Tweets\_BERT  & 0.760 &  0.669  &  N/A & N/A \\
    \hline
    Des + Loc & 0.733 & 0.693 & N/A & N/A \\
    \hline
    Des + Loc + Twt  & 0.760 & 0.725 & N/A & N/A \\
    \hline
    Des + Loc + Net & 0.775 & 0.723 & N/A & N/A  \\
    \hline
    \textbf{YUN} & \textbf{0.790} & \textbf{0.742} & N/A & N/A \\
    \hline
    GRUEmo  & N/A & N/A &  \textbf{0.931} & \textbf{0.90} \\
    \hline
    \textbf{BiLSTMAttEmo} & N/A & N/A & 0.923 & 0.891 \\
    \hline
    \end{tabular}%
  \label{tab:resultModel}%
\end{table*}%

In our user type detection model, we explore several combinations of features and evaluate YUN under ten different feature settings as baselines: Description, Location, Tweets, Network, Description\_BERT, Location\_BERT, Tweets\_BERT, Des + Loc, Des + Loc + Twt, Des + Loc + Net.

\subsubsection{Description} 
For user description, to represent texts we initialize word embeddings and emoji embeddings with 300-dimensional pre-trained Word2Vec and Emoji2Vec embeddings correspondingly. We forward these embeddings to Bi-LSTM with attention network to obtain representation of user description. We pass the description representation to a two-layer classifier activated by ReLU and then softmax.

\subsubsection{Location} 
To represent texts we follow same approach like user description. We forward these embeddings to LSTM network to obtain representation of user location. We feed the location representation to a two-layer classifier activated by ReLU and then softmax.

\subsubsection{Tweets} 
For tweets representation, we follow exactly same approach like user description representation.

\subsubsection{Network} 
After constructing user network, we use Node2Vec for network embedding. We forward the network embedding to a linear layer with ReLU activation function for obtaining user network representation. Then We pass the network representation to a two-layer classifier activated by ReLU and softmax respectively.

\subsubsection{Des + Loc} 
We concatenate description and location representations and pass the joint representation to a two-layer classifier activated by ReLU and then softmax.

\subsubsection{Des + Loc + Twt} 
Concatenation of description, location, and tweets representations are fed to the two-layer classifier with ReLU and softmax activation function.

\subsubsection{Des + Loc + Net} 
For this setting, we concatenate description, location, user network representations and forward the joint representation to our two-layer classifier activated by ReLU and softmax respectively.

\subsubsection{Description\_BERT}

We use pre-trained BERT with user description to fine-tune a model to get near state of the art performance in classification. BERT's input representation is constructed by summing the corresponding token, segment, and position embeddings for a given token. \textit{BertForSequenceClassification} model with an added single linear layer on top is used for the classification task of user type. As we feed input data, the entire pre-trained BERT model and the additional untrained classification layer is trained on our specific task. 

We use the final hidden state of the first token as the input. We denote the vector as $F \in \mathbb{R}^h$. Then we add a classifier whose parameter matrix is $W \in \mathbb{R}^{l \times h}$, where $l$ is the number of class label. Finally, the probability $P$ of each class label is calculated by the softmax function $P = softmax(FW^T)$.


\subsubsection{Location\_BERT}
In this case, we use pre-trained and fine-tuned \textit{BertForSequenceClassification} model with user location to classify user type. 
 
\subsubsection{Tweets\_BERT}
For Tweets\_BERT model, we use pre-trained and fine-tuned \textit{BertForSequenceClassification} model with user tweets to classify our tasks similar as Description\_BERT baseline model.  

\subsection{Hyperparameter Settings}
\label{subsec:6.2}

For all the models except the BERT fine-tuned models, we perform grid hyperparameter search on the validation set using early stopping. For learning rate, we explore values $0.005, 0.01, 0.05, 0.1, 0.5$; and for $L^2$ regularization, values $0, 10^{-4}, 10^{-3}, 10^{-2}$. We run the models total $30$ epochs and plot curves for loss and macro-avg F1 score. We set the criteria of early stopping such a way where the validation loss started to increase sharply. In the Description\_BERT, Location\_BERT, and Tweets\_BERT model, to encode our texts, for padding or truncating we decide maximum sentence length = $160, 50, 500$ respectively. We use batch size = $32$, learning rate = $2e-5$, AdamW optimizer \cite{loshchilov2017decoupled}, epsilon parameter = $1e-8$, number of epochs = $4$.

Network embeddings are trained using Node2Vec with parameters of dimension $300$, number of walks per source $10$, length of walk per source $80$, minimum count $1$, window size $10$. For users not appearing in the mention network, we set their network embedding vectors as $0$. We pass the network embedding to a linear layer of size $150$ with ReLU activation to compute network representation. Users not having location and/or description, we set the embedding vectors $0$ respectively.

A brief summary of hyperparameter settings of the models is shown in Table \ref{tab:hyperparam}.

\subsection{Results}
\label{subsec:6.3}
In our experiments, YUN attains the highest test accuracy $(79\%)$ and macro-avg F1 score $(74.2\%)$ for classifying user type. Table \ref{tab:resultModel} shows the performance comparison of models with baseline on test dataset.

For emotion detection in source task, baseline model GRUEmo achieves $93.1\%$ and $90\%$ test accuracy and macro-avg F1 score respectively (Table \ref{tab:resultModel}). In that case, our BiLSTMAttEmo has $92.3\%$ test accuracy and $89.1\%$ test macro-avg F1 score (Table \ref{tab:resultModel}). As the GRUEmo outperforms our BiLSTMAttEmo model for source emotion detection, we transfer knowledge from both model to detect emotion in our target task (emotion detection of yoga practitioners). For sanity check, we randomly sample $550$ tweets and label them manually with \textit{\{joy, love, sadness, anger, fear, surprise, ne\}} where \textit{\{ne\}} represents no emotion. We calculate accuracy and macro average $F1$-score. We notice that BiLSTMAttEmo model (accuracy = $66.1\%$, macro-avg F1 score = $26.3\%$) performs better in F1 score than GRUEmo model (accuracy = $72.6\%$, macro-avg F1 score = $13.9\%$) for detecting emotion in target task.
\begin{table}[htbp]
  \centering
  \caption{Granger causality result for lag = 5 }
    \begin{tabular}{|c|c|c|c|c|}
    \hline
    \textbf{Users} & \textbf{Feature}  & \textbf{rn}  & \textbf{kn}  & \textbf{nc}  \\ 
    \hline
    
    \multirow{2}{*} {All}  & y + h & 1663 & 7120 & 2271 \\
    \cline{2-5}
    & \textbf{y + $1^{st}$ + h} & 1447 & 4524 & 5083 \\
    \hline
    \hline
    \multirow{2}{*} {Top $10\%$} & y + h & 700  & 399 & 6 \\
    \cline{2-5}
      & \textbf{y + $1^{st}$ + h } & 546 & 551 & 8\\
    \hline
    \hline
    \multicolumn{5}{|p{30em}|}{rn: Number of users for whom we reject null hypothesis.} \\
    \multicolumn{5}{|p{30em}|}{kn: Number of users for whom we keep null hypothesis.} \\
    \multicolumn{5}{|p{30em}|}{nc: Number of users whose Granger Causality is not calculated.} \\
    \multicolumn{5}{|p{30em}|}{y + h: Yoga and happiness.} \\
    \multicolumn{5}{|p{30em}|}{y + $1^{st}$ + h: Yoga with $1^{st}$ hand experience and happiness.} \\
     \hline
    \end{tabular}%
  \label{tab:resultGC}%
\end{table}%
We run Granger causality (see \hyperref[subsec:3.1]{subsection III.A}) in each practitioner data with different lags = $1, 2, 3, 4, 5$. We choose lags by running the Granger test several times. Our null hypothesis is ``yoga does not Granger-cause happiness". We reject the null hypothesis if $p$-value is $\leq 0.05$, otherwise we accept the null hypothesis. 

In our case, there are $8813$ practitioners who tweeted about `yoga' and having positive emotion. With lag = $5$, we find that there are $1447$ practitioners for whom ``yoga Granger-causes happiness". For $4524$ practitioners we could not find causation with yoga and happiness. However, for $5083$ practitioners we did not have enough data to calculate Granger causality (Table \ref{tab:resultGC}). 

We select top $10\%$ practitioners who tweeted most about `yoga'.
After running Granger causality in each of $1105$ practitioners with $5$ lags, we observe that for $546$ practitioners, practicing yoga makes them happy. $551$ practitioners do not have causation with yoga and happiness. Due to the lack of enough data, we could not calculate Granger causality for $8$ practitioners. 

Table \ref{tab:resultGC} shows the combined results of Granger causality test on each practitioner among $15168$ as well as top $10\%$ practitioners ($1105$) based on both features such as `yoga + happiness' and `yoga with first hand experience + happiness'. The results are different for those two features because `yoga + happiness' considers all tweets containing `yoga'. It does not include practitioner's first hand experience. On the other hand, `yoga with first hand experience + happiness' feature focuses on yoga activity based on practitioner's first hand experience (details in \hyperref[subsec:4.4]{subsection IV.D}). 

To check if the overall activity and happiness of practitioners are temporally correlated, we test Granger causality with the null hypothesis ``tweeting more does not Granger-cause positive emotion”. In this case, we have two causal features (i) number of tweets (cause), (ii) number of positive emotions (effect). We calculate Granger causality (similar as equation \eqref{eq:1}) with lags = $1, 2, 3, 4, 5$ and for each lag, we notice that $p$-value is $> 0.05$. So we could not reject the null hypothesis which means the overall activity of the practitioners has no effect on happiness.

As in our work, we did not use contextualized word embedding for text; there are misclassifications among user types. Moreover, our transfer learning approach for the emotion detection of yoga practitioners is noisy. Also, data annotation is expensive. A future direction could be developing a contextualized model to predict user type using minimal supervision and applying a similar approach to detect users' emotions.

\section{Conclusion and Future Work}
\label{sec:7}
We analyze the causal relationship between \textit{practicing yoga} and \textit{being happy} by leveraging textual and temporal information of Twitter users using Granger causality. To measure two causal features such as (i) understanding user type, we propose a joint embedding attention-based neural network model by incorporating social and textual information of users, (ii) detecting emotion of specific users, we transfer knowledge from attention-based neural network model trained on a  source domain. Methodologically, there is more room for improvement, such as our transfer learning approach for emotion is noisy. In future work, we aim to develop a more nuanced way to measure the emotional state of yoga practitioners. We plan to employ the extended version of the proposed methodology in different life-style decision choices i.e. `keto diet', `veganism'.

\bibliography{happybib}{}
\bibliographystyle{ieeetr}
\end{document}